\documentclass[11pt,a4paper]{llncs}
\usepackage{amsmath}
\usepackage{amssymb}
\usepackage{graphicx}
\usepackage{marvosym}
\usepackage{url}
\usepackage{fancyhdr}
\usepackage{hyperref}
\usepackage{geometry}
\usepackage{rotating}
\usepackage{float}
\newcommand{\keywords}[1]{\par\addvspace\baselineskip
\noindent\keywordname\enspace\ignorespaces#1}

\fancypagestyle{firstpage}{\fancyhf{}
}

\begin{document}

%\mainmatter  % start of an individual contribution

% first the title is needed
\title{\LARGE{Arabic Metaphor Sentiment Classification Using Semantic Information}}

% a short form should be given in case it is too long for the running head
%\titlerunning{Lecture Notes in Computer Science: Authors' Instructions}

% the name(s) of the author(s) follow(s) next
%
% NB: Chinese authors should write their first names(s) in front of
% their surnames. This ensures that the names appear correctly in
% the running heads and the author index.
%
\author{\large{Israa Alsiyat}}
\institute{\large{School of Computing and Communications, Lancaster University, UK\\ and College of Science, Northern Border University}}

%\author{Alfred Hofmann%
%\thanks{Please note that the LNCS Editorial assumes that all authors have used
%the western naming convention, with given names preceding surnames. This determines
%the structure of the names in the running heads and the author index.}%
%\and Ursula Barth\and Ingrid Haas\and Frank Holzwarth\and\\
%Anna Kramer\and Leonie Kunz\and Christine Rei\ss\and\\
%Nicole Sator\and Erika Siebert-Cole\and Peter Stra\ss er}
%
%\authorrunning{Lecture Notes in Computer Science: Authors' Instructions}
% (feature abused for this document to repeat the title also on left hand pages)

% the affiliations are given next; don't give your e-mail address
% unless you accept that it will be published
%\institute{Springer-Verlag, Computer Science Editorial,\\
%Tiergartenstr. 17, 69121 Heidelberg, Germany\\
%\mailsa\\
%\mailsb\\
%\mailsc\\
%\url{http://www.springer.com/lncs}}

%
% NB: a more complex sample for affiliations and the mapping to the
% corresponding authors can be found in the file "llncs.dem"
% (search for the string "\mainmatter" where a contribution starts).
% "llncs.dem" accompanies the document class "llncs.cls".
%

%\toctitle{Lecture Notes in Computer Science}
%\tocauthor{Authors' Instructions}

\maketitle

\thispagestyle{firstpage}
\begin{abstract}
In this paper, I discuss the testing of the Arabic Metaphor Corpus (AMC) \cite{alsiyat2023amc} using newly designed automatic tools for sentiment classification for AMC based on semantic tags. The tool incorporates semantic emotional tags for sentiment classification. I evaluate the tool using standard methods, which are F-score, recall, and precision. The method is to show the impact of Arabic online metaphors on sentiment through the newly designed tools. To the best of our knowledge, this is the first approach to conduct sentiment classification for Arabic metaphors using semantic tags to find the impact of the metaphor.
\end{abstract}

%\begin{abstract}
%The abstract should summarize the contents of the paper and should
%contain at least 70 and at most 150 words. It should be written using the
%\emph{abstract} environment.
%\keywords{We would like to encourage you to list your keywords within
%the abstract section}
%\end{abstract}

\section{Introduction}
To the best of our knowledge, there are no existing tools specifically developed for Arabic metaphor identification in the context of sentiment analysis. Identifying Arabic metaphors requires pre-annotated data, and in the absence of pre-annotation, a substantial corpus of Arabic metaphors would be necessary to train advanced machine learning algorithms for automatic identification. So, I am using the Arabic Metaphor Corpus (AMC) \cite{alsiyat2023amc}. 

In terms of the Arabic metaphor, the very recent study conducted for Arabic metaphor identification with pre-annotated data without integrating sentiment classification is \cite{abugharsa2022metaphor}. \cite{abugharsa2022metaphor} conducted a binary classification for metaphor identification with no sentiment following a certain method used to adapt the LSTM (Long-Short Term Memory) method of identification. This method is accurate for the Libyan dialect for poems, which could be not applicable for the online Arabic metaphor that has a different structure. However, our method is to eliminate human intervention for annotation to identify sentiment with metaphor (AMC) using the Arabic Semantic Tagger \cite{arasas} \ref{AMC}.

The reliable existing lexicon resources, annotated corpora and algorithms devised for English facilitate the process of metaphor identification, such as BNC \cite{BNC} British National Corpus ,WordNet \cite{miller1995wordnet}, VerbNet \cite{schuler2005verbnet} and TreeBank \cite{marcus1993buildingTreeBank}. As an example of the existing algorithms, Word Sense Disambiguation (WSD) contributes to identifying sentiment and metaphor in English in \cite{rentoumi2012sentiment}, but although WSD has many applications for Arabic, it has none for metaphor \cite{HadniWSDArabic}. 

In this paper, I present the first Arabic sentiment classification with metaphor and semantic information.

\subsection{The Arabic metaphor Corpus}
\label{AMC}
The Arabic Metaphor Corpus (AMC) consists of 1,000 Arabic book reviews that have been manually annotated for metaphor terms, sentiment analysis, and gold standards. The sentiment annotations encompass both the overall sentiment of each review and the sentiment specific to the identified metaphors. In addition, the corpus includes automatically generated semantic annotations derived from the Arabic Semantic Tagger (AraSAS) \cite{arasas}. The AMC dataset is available in Excel format and can be accessed publicly via the following link: \href{https://github.com/IsraaMousa/IsraaArabicMetaphor/find/main}{AMC}.

\subsection{Arabic Semantic Tagger (AraSAS)}
AraSAS is an online tool developed for annotating Arabic text with semantic tags \cite{arasas}. This tool provides a user-friendly interface that supports Arabic text input, with a processing capacity of up to 100,000 words. Users can choose from three tagging formats: horizontal, vertical, and XML. The tool tags the input text based on a predefined set of subcategories, which are outlined in the input guidelines available at \href{https://arasas.herokuapp.com/}{AraSAS}. However, it does not support the annotation of certain special characters, such as the asterisk (*).

In terms of sentiment classification, AraSAS assigns positive sentiment to tags marked with (+), (++), and (+++), and negative sentiment to tags with (-), (--), and (---). The number of signs indicates the degree of intensity within the sentiment, but the tool does not provide explicit sentiment strength classifications. Importantly, AraSAS does not have built-in functionality to classify the overall sentiment of the tagged text.

For the purposes of this study, AraSAS was utilized to annotate the reviews within the AMC. Due to the tool’s limitation of processing a maximum of 100 reviews at a time, the AMC was divided into batches of 100 reviews each. These batches were stored in Excel files and processed independently. To prevent erroneous segmentation of text, punctuation marks such as full stops and exclamation marks, which AraSAS treats as line breaks, were substituted with English letters prior to annotation.

\section{Experiment}
This section is divided into multiple steps: starting with the two tools for sentiment classification in regard to the metaphor section only and in regard to the review as well as considering the annotation for metaphor. Then the tool optimized to classify the sentiment to cover all semantic tags. Mean the optimization steps will be discussed to show the different results of the classification by observing the classification of each stage.

\subsection{Sentiment classification tool without metaphor for overall sentiment}
The function written to classify the sentiment was based on the semantically tagged ACM dataset. The tagged dataset was turned into a data frame using Python. The function was used to classify sentiment by counting the number of positive and negative emotional tags (E tags only). Each review was subjected to this function to calculate the sentiment score. The sentiment score and the polarity were regarded as a dictionary and served as an argument to the count function, which were the emotional tags without any sign of polarity.

The function was designed to not calculate the neutral E-tags in order to avoid the redundancy of counting similar tags. For example, if E1 is added to the list, the counter regards E1 and E1+ as a redundancy or as the same. In addition, there were sentences in which the numbers of negative and positive tags were equal, and in some reviews,the numbers of neutral, positive, and negative tags were all equal. This means that the review had the same amount of polarities. For example, when the emotional tags counted as positive, negative, and neutral were equal, the neutral ones were removed to avoid any complications because my aim was to test the differences between the methods and to test the AMC using this method. I therefore deleted the neutral tags from the list of emotional tags so that only the negative and positive tags were defined beforehand. I will evaluate these methods further in the Conclusion chapter as suggestions for further research, because for better sentiment classification in regards to metaphor, all the polarity signs should be considered in the classification.

The calculation of the sentiment score was based on a set of conditions after linking the emotional tag with the polarity. So for a positive emotional tag (item) from the pre-defined list, I added 0.5 to the count, and the opposite for the negative emotional tags, by subtracting 0.5. For a double positive E++ the function adds 1 to the count, and for double negative signs (--) the function subtracts 1 from the count. For triple positive emotional tags (E+++) the function adds 1.5, and for triple negatives (E---) it subtracts 1.5. These operations were carried out for each review, and in this way the summation of the sentiment score was calculated. Finally, sentiment was classified based on the sentiment score. If the sentiment score is greater than zero, the polarity is positive; if it is less than zero, the polarity is negative. Otherwise, the polarity is neutral.
\begin{figure}[htbp]
    \centering    
    \includegraphics[width=20cm]{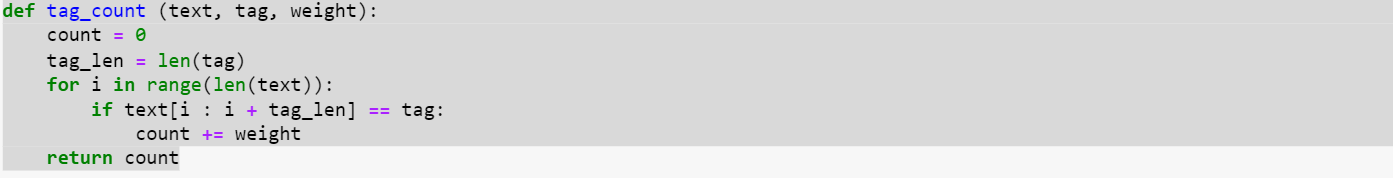}
    \caption{Sentiment classification using AraSAS} %\citep{alsiyat2020metaphorical}}
    \label{fig:count function for AraSAS tags}
\end{figure}
%\begin{figure}[hbt!]
\begin{figure}[htbp]
    \centering    
    \includegraphics[width=\textwidth]{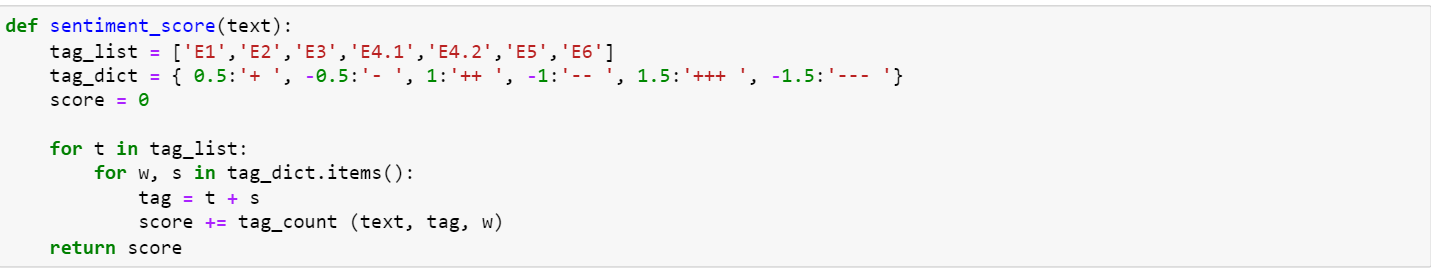}
    \caption{Sentiment classification using AraSAS} % \citep{alsiyat2020metaphorical}
    \label{fig:SentimentClassificationAraSAS}
\end{figure}
%\begin{figure}
\begin{figure}[htbp]
    \centering
    \includegraphics[width=20cm]{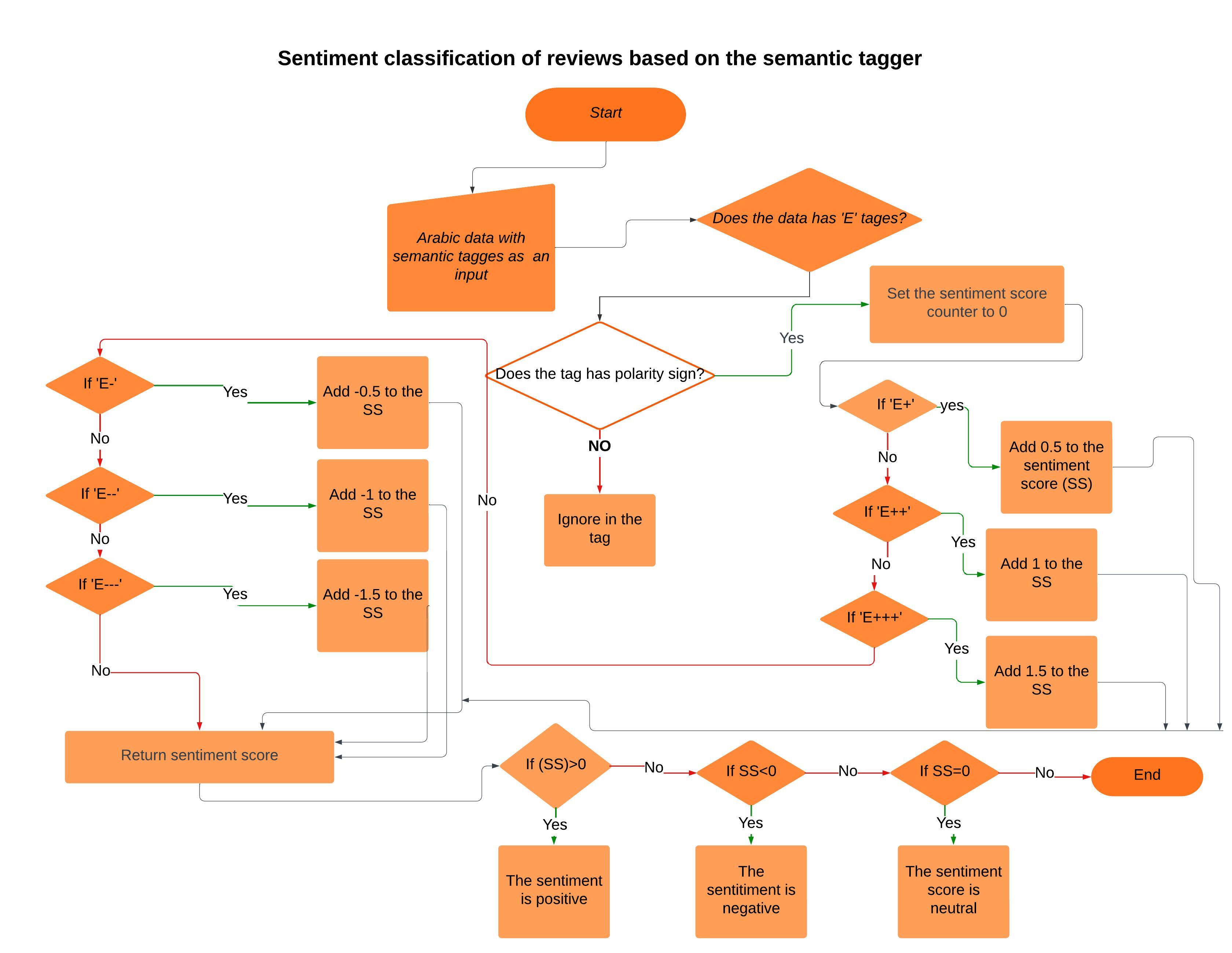}
    \caption{Sentiment detection tool} %\citep{alsiyat2020metaphorical}}
    \label{Sentiment detection tool}
%\end{figure}
\end{figure}
%\begin{figure}
\clearpage

\subsection{sentiment classification tool considering metaphor}
In this method, for calculating the sentiment score for a review, a score of 2 is added to the sentiment score if the polarity is positive and subtracted 2 if it is negative. If it is neutral or null, return the sentiment score of zero. I regarded this as metaphor sentiment classification with a semantic tagger as it followed the AraSAS \cite{arasas} sentiment score tagging. 

The flow chart in Figure \ref{Sentiment detection tool} shows the initial framework designed to classify sentiment using the tagged data from the Arabic semantic tagger. The flow chart explains the code that performs the overall sentiment classification. As previously explained, the E-tag was the main tag for tagging emotion in the semantic tagger.

Another program was designed to detect review sentiment based on the sentiment score given by the manual annotation for each review. The classification was based on the sentiment score. 

As has already been discussed, the classification could be more precise if it is performed based on the numbers and types of the polarity sign for the E-tags. For example, before assigning a sentiment score to an E-tag, all E-tags with each polarity sign could be counted and the total could be assessed against a set of conditions in order to detect the overall polarity. Then the sentiment score can be assigned based on the labelled data. This means that if the polarity is negative, the sentiment will be -1, if it is positive 1 and if it is neutral 0. However, this method did not consider the degree of the positivity and negativity for each tag, which I considered in our classification in the suggestion. Due to the cases which I encountered during the classification, such as the equality of the E-tags and polarity signs, the work needs to be adapted to fit all cases and the metaphor detection, which would have exceeded the time frame set for this project. The sentiment classification was therefore implemented using the sentiment scores only.

\begin{figure}[hbt!]
    \centering    
    \includegraphics[width=8cm]{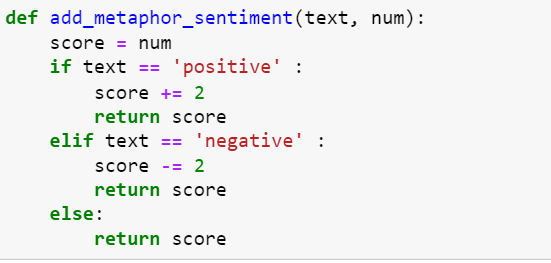}
    \caption{Sentiment classification and sentiment score for metaphor}
    \label{fig:Sentiment classification for metaphor}
\end{figure}
\clearpage
\begin{figure}[ht]
    \centering
    \includegraphics[width=20cm]{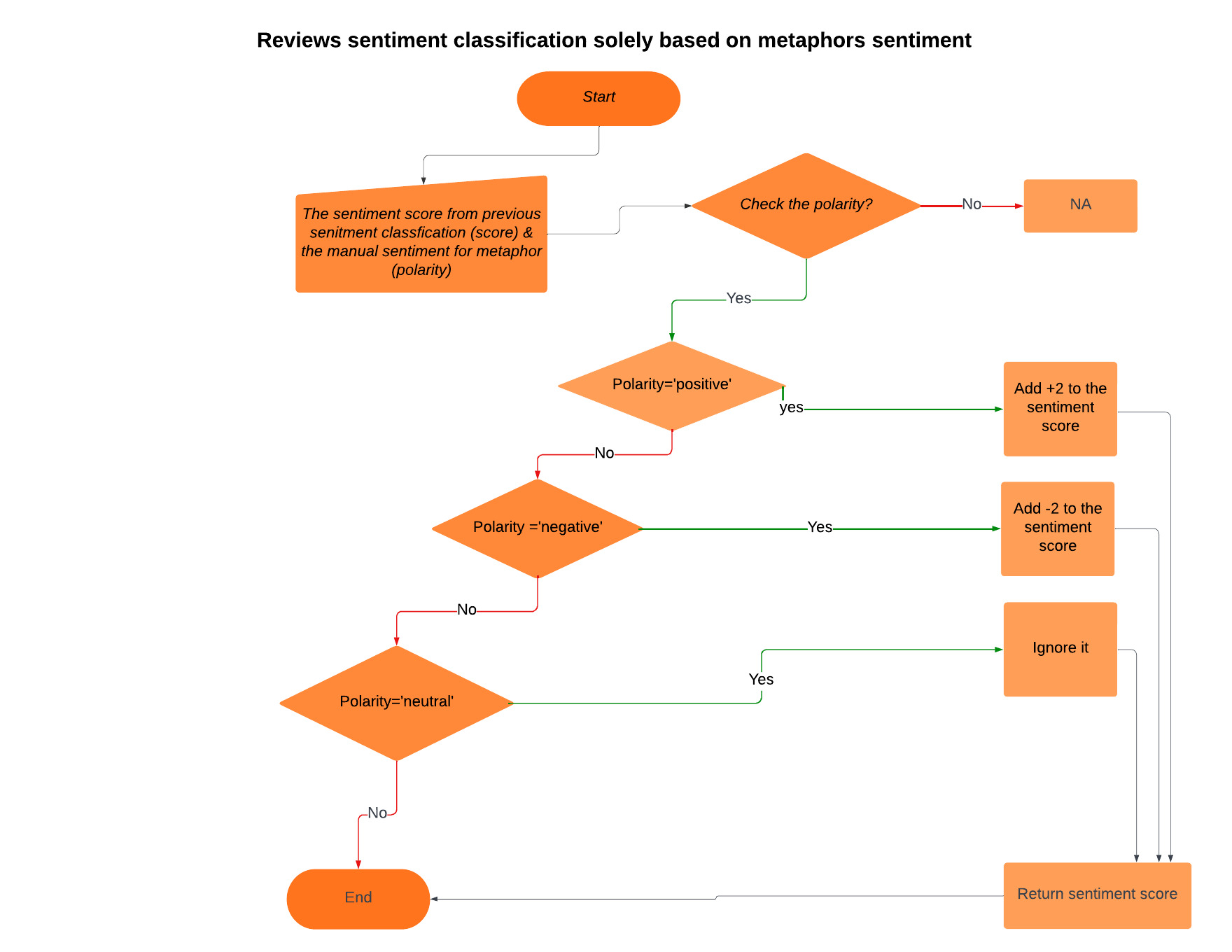}
    \caption{Metaphor sentiment classification tool} %\citep{alsiyat2020metaphorical}}
    \label{Metaphor sentiment tool}
\end{figure}
%\end{sidewaysfigure}
\clearpage

\section{Evaluation}
In this section, I assess the methods for identifying the sentiment of reviews in regard to metaphor using semantic information. The performances of these methods were assessed using the standard measurements of precision, recall and F-score. The calculation of statistics was done automatically using a Python code shown in Figure \ref{fig:evaluation python code}. 
The classification results obtained by the methods were assessed using the Python code with equations \ref{equ:F-score}, \ref{equ:Precision} and \ref{equ:recall}. Then the F-score was used to evaluate the performances of the methods.

\begin{equation}
\small
\label{equ:Precision}
Precision = \frac{True positive}{True positive + False positive}
\end{equation}

\begin{equation}
\small
\label{equ:recall}
Recall = \frac{True positive}{True positive + False negative}
\end{equation}

\begin{equation}
\small
\label{equ:F-score}
F-Score= \frac{2*precision*recall}{precision + recall}
\end{equation}

The Arabic semantic tagger is not designed to identify sentiment. So I designed a program to classify sentiment based on basic sentiment represented by the emotional tags. The following methods are for sentiment annotation in regard to metaphors which was used to analyze the performance of each model based on the automatic tools explained above.
The formulas \ref{equ:Precision}\ref{equ:recall} are not applied manually to the multi-class recall and precision calculation. The annotation data cases from each model have to be extracted to calculate the recall and precision automatically. In detail, the cases of the methods from the gold standard and one of the models had to be extracted prior to the calculation. For example, by counting the number of negative agreements between the gold standard and one of the models, and repeat it for all classes. Then the equations \ref{equ:Precision} and \ref{equ:recall} were applied to the extracted tables for the methods. In addition, the calculation for the precision, recall and F-score are applied for all other categories of reviews. For example, the recall and the precision will be applied to all the positive review categories for the all methods.

Each method discussed below has the table containing the sentiment classification method and the method name. The sentiment classification methods have the review category and number of reviews. The review category has the reviews lengths ranges. The number of reviews column has the count number of the reviews for the correspond category. In addition, to the method name contains the calculations numbers of the standard equations \ref{equ:F-score}, \ref{equ:Precision} and \ref{equ:recall}.

\subsection{Sentiment classification tool without metaphor for overall sentiment}
\begin{table}[hbt!]
\centering
\begin{tabular}{|cc|ccc|}
\hline
\multicolumn{2}{|c|}{\textbf{Sentiment classification}} & \multicolumn{3}{c|}{\textbf{\begin{tabular}[c]{@{}c@{}}Sentence sentiment classification\\ based on gold standard\end{tabular}}} \\ \hline
\multicolumn{1}{|c|}{\textbf{Review categories}} & \textbf{Num. of revi.} & \multicolumn{1}{c|}{\textbf{Precision}} & \multicolumn{1}{c|}{\textbf{Recall}} & \textbf{F-score} \\ \hline
\multicolumn{1}{|c|}{\textbf{All reviews}} & 1000 & \multicolumn{1}{c|}{0.641} & \multicolumn{1}{c|}{0.487} & 0.526 \\ \hline
\multicolumn{1}{|c|}{\textbf{Positive   reviews}} & 702 & \multicolumn{1}{c|}{0.831} & \multicolumn{1}{c|}{0.520} & 0.640 \\ \hline
\multicolumn{1}{|c|}{\textbf{Negative   reviews}} & 171 & \multicolumn{1}{c|}{0.263} & \multicolumn{1}{c|}{0.626} & 0.370 \\ \hline
\multicolumn{1}{|c|}{\textbf{Neutral   reviews}} & 127 & \multicolumn{1}{c|}{0.097} & \multicolumn{1}{c|}{0.118} & 0.107 \\ \hline
\multicolumn{1}{|c|}{\textgreater{}=1000 tks} & 5 & \multicolumn{1}{c|}{0.000} & \multicolumn{1}{c|}{0.000} & 0.000 \\ \hline
\multicolumn{1}{|c|}{999 tks $\sim$500   tks} & 32 & \multicolumn{1}{c|}{0.686} & \multicolumn{1}{c|}{0.375} & 0.447 \\ \hline
\multicolumn{1}{|c|}{499 tks $\sim$100   tks} & 268 & \multicolumn{1}{c|}{0.617} & \multicolumn{1}{c|}{0.369} & 0.426 \\ \hline
\multicolumn{1}{|c|}{99 tks $\sim$90   tks} & 24 & \multicolumn{1}{c|}{0.747} & \multicolumn{1}{c|}{0.500} & 0.549 \\ \hline
\multicolumn{1}{|c|}{79 tks $\sim$70   tks} & 30 & \multicolumn{1}{c|}{0.783} & \multicolumn{1}{c|}{0.633} & 0.660 \\ \hline
\multicolumn{1}{|c|}{69 tks $\sim$60   tks} & 31 & \multicolumn{1}{c|}{0.882} & \multicolumn{1}{c|}{0.452} & 0.582 \\ \hline
\multicolumn{1}{|c|}{59 tks $\sim$50   tks} & 57 & \multicolumn{1}{c|}{0.651} & \multicolumn{1}{c|}{0.456} & 0.512 \\ \hline
\multicolumn{1}{|c|}{49 tks $\sim$40   tks} & 53 & \multicolumn{1}{c|}{0.548} & \multicolumn{1}{c|}{0.472} & 0.473 \\ \hline
\multicolumn{1}{|c|}{39 tks $\sim$30   tks} & 70 & \multicolumn{1}{c|}{0.679} & \multicolumn{1}{c|}{0.500} & 0.548 \\ \hline
\multicolumn{1}{|c|}{29 tks $\sim$20   tks} & 99 & \multicolumn{1}{c|}{0.620} & \multicolumn{1}{c|}{0.556} & 0.577 \\ \hline
\multicolumn{1}{|c|}{19 tks $\sim$10   tks} & 149 & \multicolumn{1}{c|}{0.599} & \multicolumn{1}{c|}{0.557} & 0.568 \\ \hline
\multicolumn{1}{|c|}{9 tks $\sim$5 tks} & 90 & \multicolumn{1}{c|}{0.689} & \multicolumn{1}{c|}{0.600} & 0.621 \\ \hline
\multicolumn{1}{|c|}{4 tks $\sim$1 tks} & 63 & \multicolumn{1}{c|}{0.744} & \multicolumn{1}{c|}{0.603} & 0.661 \\ \hline
\end{tabular}
\caption{GS annotation method measurement} 
\end{table}

This method was explained in detail above. Here, I shall explain the evaluation method and the resulting table to analyze the method’s performance based on the F-score. The gold standard was used to compare and measure the performances of the other methods. A fifth column was added to the csv table and the Python code was used to evaluate the other method. The gold standard column acted as a sentiment score according to a set of conditions by using equations to calculate the standard measurement.

The highest F-score for this method was found to be 0.661, which was also the same as was found for the review category of between 1 and 4 tokens. The annotation for this category showed the highest scores for the two methods, which indicates that the sentiment scores for the short metaphor reviews were same as the gold standard rather than the long reviews. The highest F-scores means that the annotation was close to the gold standard annotation whereas the lowest F-score was 0.0 for the less than or equal to 1000 tokens category.

\subsection{sentiment classification tool considering metaphor}
\begin{table}[hbt!]
\centering
\begin{tabular}{|cc|ccc|}
\hline
\multicolumn{2}{|c|}{\textbf{Sentiment classification}} & \multicolumn{3}{c|}{\textbf{\begin{tabular}[c]{@{}c@{}}Sent. senti. classification based on both \\ semantic tags and\\metaphor senti. Info.\end{tabular}}} \\ \hline
\multicolumn{1}{|c|}{\textbf{Review categories}} & \textbf{Num. of rev.} & \multicolumn{1}{c|}{\textbf{Precision}} & \multicolumn{1}{c|}{\textbf{Recall}} & \textbf{F-score} \\ \hline
\multicolumn{1}{|c|}{\textbf{All reviews}} & 1000 & \multicolumn{1}{c|}{0.564} & \multicolumn{1}{c|}{0.305} & 0.351 \\ \hline
\multicolumn{1}{|c|}{\textbf{Positive   reviews}} & 702 & \multicolumn{1}{c|}{0.735} & \multicolumn{1}{c|}{0.285} & 0.411 \\ \hline
\multicolumn{1}{|c|}{\textbf{Negative   reviews}} & 171 & \multicolumn{1}{c|}{0.190} & \multicolumn{1}{c|}{0.281} & 0.226 \\ \hline
\multicolumn{1}{|c|}{\textbf{Neutral   reviews}} & 127 & \multicolumn{1}{c|}{0.120} & \multicolumn{1}{c|}{0.449} & 0.189 \\ \hline
\multicolumn{1}{|c|}{\textgreater{}=1000 tks} & 5 & \multicolumn{1}{c|}{1.000} & \multicolumn{1}{c|}{0.200} & 0.333 \\ \hline
\multicolumn{1}{|c|}{999 tks $\sim$500   tks} & 32 & \multicolumn{1}{c|}{0.637} & \multicolumn{1}{c|}{0.500} & 0.526 \\ \hline
\multicolumn{1}{|c|}{499 tks $\sim$100   tks} & 268 & \multicolumn{1}{c|}{0.554} & \multicolumn{1}{c|}{0.354} & 0.415 \\ \hline
\multicolumn{1}{|c|}{99 tks $\sim$90   tks} & 24 & \multicolumn{1}{c|}{0.582} & \multicolumn{1}{c|}{0.458} & 0.502 \\ \hline
\multicolumn{1}{|c|}{79 tks $\sim$70   tks} & 30 & \multicolumn{1}{c|}{0.580} & \multicolumn{1}{c|}{0.467} & 0.513 \\ \hline
\multicolumn{1}{|c|}{69 tks $\sim$60   tks} & 31 & \multicolumn{1}{c|}{0.729} & \multicolumn{1}{c|}{0.355} & 0.437 \\ \hline
\multicolumn{1}{|c|}{59 tks $\sim$50   tks} & 57 & \multicolumn{1}{c|}{0.529} & \multicolumn{1}{c|}{0.298} & 0.326 \\ \hline
\multicolumn{1}{|c|}{49 tks $\sim$40   tks} & 53 & \multicolumn{1}{c|}{0.577} & \multicolumn{1}{c|}{0.340} & 0.359 \\ \hline
\multicolumn{1}{|c|}{39 tks $\sim$30   tks} & 70 & \multicolumn{1}{c|}{0.601} & \multicolumn{1}{c|}{0.243} & 0.332 \\ \hline
\multicolumn{1}{|c|}{29 tks $\sim$20   tks} & 99 & \multicolumn{1}{c|}{0.561} & \multicolumn{1}{c|}{0.323} & 0.359 \\ \hline
\multicolumn{1}{|c|}{19 tks $\sim$10   tks} & 149 & \multicolumn{1}{c|}{0.599} & \multicolumn{1}{c|}{0.275} & 0.271 \\ \hline
\multicolumn{1}{|c|}{9 tks $\sim$5 tks} & 90 & \multicolumn{1}{c|}{0.414} & \multicolumn{1}{c|}{0.144} & 0.099 \\ \hline
\multicolumn{1}{|c|}{4 tks $\sim$1 tks} & 63 & \multicolumn{1}{c|}{0.878} & \multicolumn{1}{c|}{0.159} & 0.170 \\ \hline
\end{tabular}
\caption{Semantic tagger annotation method measurement with metaphor}
\label{table:both}
\end{table}
This method was divided into two steps: the first to detect the sentiment based on the semantically tagged AMC and the second to convert the GS for the metaphor sentiment annotation into a sentiment score. The first step had to meet a set of conditions and functions in order to classify the sentiment (see Figure \ref{fig:count function for AraSAS tags} and Figure \ref{fig:SentimentClassificationAraSAS}. This method is a combination of the semantic tagger and the metaphor. The metaphor function \ref{fig:Sentiment classification for metaphor} provided the sentiment score annotation from the AraSAS, which has sentiment scores calculated based on the semantic tags polarity scores. The polarity scores were specified on the semantic tagger's sub-categories.

For the GS of sentiment metaphor method evaluation \ref{fig:Sentiment classification for metaphor}, the precision, recall, and F-score were automatically calculated using the metaphor gold standard annotation with the sentiment classification using the Arabic semantic tagger.

In this method, the metaphor sentiment in the function is to be represented as text and the sentiment score as number parameters. The sentiment as a text parameter meets a set of conditions after assigning the sentiment with the sentiment score. The function's algorithm adds 2 if the sentiment is positive and subtracts 2 if it is negative. The function consists of a series of condition statements, but it classifies metaphor sentiment based only on the metaphor gold standard annotation and not on detecting the metaphors in the text. There was no metaphor detection. After the classification, the new classification column was added to the classification table. 
The table below shows the values calculated from the standard measurements, which are precision, recall and F-score for all categories. This method has lower F-scores than the other methods. The evaluation was done automatically using the Python code. The highest F-score was 0.52 for between 500 and 999 tokens and the lowest was 0.09 for the category between 5 and 9 tokens in the reviews. These F-score results are logical as the metaphor annotations were compared with the overall sentiment gold standard during the calculations, which were completely different types of text.

\section{F-scores Comparison}
The table shows the highest F-scores for each category extracted from the evaluation tables for the two tools. The table contains the categories of each review length extracted from the first table during the evaluation as I found the review length affects the sentiment classification. In addition, there are two columns of the highest F-scores of each category for the two tools. The last column contains the highest F-scores between the two tools for each category.

The highest F-score is 0.9 from the second tool for the review length between 5 and 1; this result confirms that the tool predicts accurately with a few number of words/tokens. In addition, the lowest F-score has 0.33 for the tokens above or equal to 1000, which means the tool's prediction is not highly accurate with the high number of tokens.

This means that the sentiment with metaphor is affected by the other factors mentioned in the review, which need further development for tools that contain more features to capture other aspects of the text with metaphor.

\begin{table}[H]
\centering
\begin{tabular}{|cccc|}
\hline
\multicolumn{4}{|c|}{Compare the F-scores of the Two Tools} \\ \hline
\multicolumn{1}{|c|}{Categories} &
  \multicolumn{1}{c|}{\textbf{F-score for first tool}} &
  \multicolumn{1}{c|}{\textbf{F-score for the 2nd tool}} &
  \begin{tabular}[c]{@{}c@{}}Highest F-score \\ for each category\end{tabular} \\ \hline
\multicolumn{1}{|c|}{All   reviews} &
  \multicolumn{1}{c|}{\textbf{0.3510625}} &
  \multicolumn{1}{c|}{0.70831217} &
  0.70831217 \\ \hline
\multicolumn{1}{|c|}{Negative reviews} &
  \multicolumn{1}{c|}{0.22641509} &
  \multicolumn{1}{c|}{0.59166667} &
  0.59166667 \\ \hline
\multicolumn{1}{|c|}{Neutral reviews} &
  \multicolumn{1}{c|}{0.18936877} &
  \multicolumn{1}{c|}{0.17708333} &
  0.18936877 \\ \hline
\multicolumn{1}{|c|}{\textless 10 tokens and \textgreater{}=5 tokens} &
  \multicolumn{1}{c|}{0.09908113} &
  \multicolumn{1}{c|}{0.81952828} &
  0.81952828 \\ \hline
\multicolumn{1}{|c|}{\textless 100 tokens and \textgreater{}=90 tokens} &
  \multicolumn{1}{c|}{0.50169221} &
  \multicolumn{1}{c|}{0.70714286} &
  0.70714286 \\ \hline
\multicolumn{1}{|c|}{\textless 1000 tokens and \textgreater{}=500 tokens} &
  \multicolumn{1}{c|}{0.52604908} &
  \multicolumn{1}{c|}{0.61724624} &
  0.61724624 \\ \hline
\multicolumn{1}{|c|}{\textless 20 tokens and \textgreater{}=10 tokens} &
  \multicolumn{1}{c|}{0.27081778} &
  \multicolumn{1}{c|}{0.7734014} &
  0.7734014 \\ \hline
\multicolumn{1}{|c|}{\textless 30 tokens and \textgreater{}=20 tokens} &
  \multicolumn{1}{c|}{0.35871212} &
  \multicolumn{1}{c|}{0.72564525} &
  0.72564525 \\ \hline
\multicolumn{1}{|c|}{\textless 40 tokens and \textgreater{}=30 tokens} &
  \multicolumn{1}{c|}{0.33173353} &
  \multicolumn{1}{c|}{0.68374172} &
  0.68374172 \\ \hline
\multicolumn{1}{|c|}{\textless 5 tokens and \textgreater{}=1 tokens} &
  \multicolumn{1}{c|}{0.1704478} &
  \multicolumn{1}{c|}{0.95825325} &
  0.95825325 \\ \hline
\multicolumn{1}{|c|}{\textless 50 tokens and \textgreater{}=40 tokens} &
  \multicolumn{1}{c|}{0.35893024} &
  \multicolumn{1}{c|}{0.69212357} &
  0.69212357 \\ \hline
\multicolumn{1}{|c|}{\textless 500 tokens and \textgreater{}=100 tokens} &
  \multicolumn{1}{c|}{0.41484196} &
  \multicolumn{1}{c|}{0.59772096} &
  0.59772096 \\ \hline
\multicolumn{1}{|c|}{\textless 60 tokens and \textgreater{}=50 tokens} &
  \multicolumn{1}{c|}{0.32623225} &
  \multicolumn{1}{c|}{0.67541087} &
  0.67541087 \\ \hline
\multicolumn{1}{|c|}{\textless 70 tokens and \textgreater{}=60 tokens} &
  \multicolumn{1}{c|}{0.4365268} &
  \multicolumn{1}{c|}{0.79286205} &
  0.79286205 \\ \hline
\multicolumn{1}{|c|}{\textless 80 tokens and \textgreater{}=70 tokens} &
  \multicolumn{1}{c|}{0.51313131} &
  \multicolumn{1}{c|}{0.7047619} &
  0.7047619 \\ \hline
\multicolumn{1}{|c|}{\textgreater{}=1000 tokens} &
  \multicolumn{1}{c|}{0.33333333} &
  \multicolumn{1}{c|}{0.33333333} &
  0.33333333 \\ \hline
\multicolumn{1}{|c|}{Positive reviews} &
  \multicolumn{1}{c|}{0.41067762} &
  \multicolumn{1}{c|}{0.83283133} &
  0.83283133 \\ \hline
\end{tabular}
\end{table}

\section{Conclusion}
 In this paper, I discussed the newly designed two Arabic sentiment tools for metaphor (AMC)\cite{alsiyat2023amc} using semantic information. Those two tools are promising for Arabic sentiment classification with metaphor using the semantic information. The semantic tags can be used to even test the existence of metaphor, which is an alternative solution for pre-annotation data for Arabic metaphor detection.
 
 The tools were evaluated using the standard measurements, which are precision, recall and F-score. At the end, I discussed the impact of metaphor on sentiment by comparing the F-scores of the two tools. The impact was discussed through the distinctive F-score results of the two tools. In addition, the review lengths were discussed as another factor affecting the sentiment classification for metaphor.

\section{Acknowledgments}
The author extends their appreciation to the Deanship of Scientific Research at the Department of Computing, College of Science, Northern Border University, Arar, KSA for funding this research work through the project number "NBU-SAFIR-2025". And the appreciation extends as well to the Royal Embassy of Saudi Arabia, Cultural and Lancaster University, UK, which involves this work.

\vspace{2cm}

\section*{Authors}
\noindent {\bf Israa Alsiyat} received a Master's degree in computer science from Bowie State University, USA, and she obtained a Bachelor's degree in computer science from Aljouf University. Currently, she is pursuing her PhD in Computer Science from Lancaster University, UK. Her research is The Impact of Online Metaphors
on Automatic Arabic Sentiment.\\

\end{document}